# Comparison of deep learning and conventional methods for disease onset prediction

Luis H. John, MSc[1], Chungsoo Kim, PhD[2], Jan A. Kors, PhD[1], Junhyuk Chang, PharmD[3], Hannah Morgan-Cooper, MSc[4], Priya Desai, MSc[4], Chao Pang, PhD[5], Peter R. Rijnbeek, PhD[1], Jenna M. Reps, PhD[1,6], Egill A. Fridgeirsson, PhD[1]

[1]Department of Medical Informatics, Erasmus University Medical Center, Rotterdam, The Netherlands

[2]Section of Cardiovascular Medicine, Department of Internal Medicine, Yale School of Medicine, New Haven, CT, United States

[3]Department of Biomedical Informatics, Ajou University Graduate School of Medicine, Suwon, Republic of Korea

[4]Stanford School of Medicine and Stanford Health Care, Palo Alto, CA, United States

[5]Department of Biomedical Informatics, Columbia University Irving Medical Center, New York, NY, United States

[6]Janssen Research and Development, Titusville, NJ, United States

## Contact Information

Luis H. John (corresponding author): l.john@erasmusmc.nl; +31 10 704 44 81; Dr. Molewaterplein 40, 3015 GD Rotterdam, The Netherlands

Chungsoo Kim: chungsoo@ohdsi.org

Jan A. Kors: j.kors@erasmusmc.nl

Junhyuk Chang: wkd9504@ajou.ac.kr

Hannah Morgan-Cooper: hannahmc@stanford.edu

Priya Desai: prd@stanford.edu

Chao Pang: cp3016@cumc.columbia.edu

Peter R. Rijnbeek: p.rijnbeek@erasmusmc.nl

Jenna M. Reps: jreps@its.jnj.com

Egill A. Fridgeirsson: e.fridgeirsson@erasmusmc.nl

## Research in context

### Evidence before this study

Conventional methods like logistic regression and gradient boosting have been widely used for their robustness and interpretability in disease onset prediction. Deep learning methods have faced challenges with structured observational healthcare data due to data sparsity and high dimensionality. Recently, adaptions of ResNet and Transformer models have become available, leading to renewed interest in their application to tabular data.

### Added value of this study

This study is the first to perform a large-scale evaluation of deep learning and conventional prediction methods across 11 diverse international databases. The focus is on external validation to evaluate model transportability, which is an aspect not commonly investigated for deep learning. By focusing on clinically relevant outcomes including lung cancer, dementia, and bipolar disorder, this study provides generalizable insights. It highlights the comparative discrimination and calibration performance, and scalability of these methods, emphasizing the strengths and limitations of each method for disease onset prediction.

**Implications of all the available evidence**

Our findings suggest that conventional models like logistic regression and gradient boosting remain highly effective, particularly with smaller datasets and during external validation. This suggests these methods should continue to be used in clinical practice for their reliability and ease of use. While deep learning shows potential, more research and development are needed to optimize these models for structured observational data. Future work should also focus on finding new strategies to exploit the full capabilities of deep learning methods for disease onset prediction.

**Summary**

**Background:** Conventional prediction methods such as logistic regression and gradient boosting have been widely utilized for disease onset prediction for their reliability and interpretability. Deep learning methods promise enhanced prediction performance by extracting complex patterns from clinical data, but face challenges like data sparsity and high dimensionality.

**Methods:** This study compares conventional and deep learning approaches to predict lung cancer, dementia, and bipolar disorder using observational data from eleven databases from North America, Europe, and Asia. Models were developed using logistic regression, gradient boosting, ResNet, and Transformer, and validated both internally and externally across the data sources. Discrimination performance was assessed using the area under the receiver operating characteristic curve (AUROC), and calibration was evaluated using $E_{avg}$.

**Findings:** Across eleven datasets, conventional methods (logistic regression and XGBoost) generally outperformed deep learning methods (ResNet and Transformer) in terms of discrimination performance, particularly during external validation, highlighting their better transportability. Learning curve analysis suggested that deep learning models require substantially larger datasets to reach the same performance levels as conventional methods. Calibration performance was also better for conventional methods, with ResNet showing the poorest calibration among all models.

**Interpretation:** Despite the potential of deep learning models to capture complex patterns in structured observational healthcare data, conventional models remain highly competitive for disease onset prediction, especially in scenarios involving smaller datasets and if lengthy training times need to be avoided. The study underscores the need for future research focused on optimizing deep learning models to handle the sparsity, high dimensionality, and heterogeneity inherent in healthcare datasets, and find new strategies to exploit the full capabilities of deep learning methods.

**Funding:** This project has received funding from the Innovative Medicines Initiative 2 Joint Undertaking (JU) under grant agreement No. 806968. The JU receives support from the European Union's Horizon 2020 research and innovation programme and EFPIA.

# 1 Introduction

Identifying individuals at high risk of disease at an early stage allows for improved care and risk-factor targeted intervention. Prognostic models can provide such risk estimates, aiding healthcare providers in making informed clinical decisions which can ultimately improve patient outcomes.(1, 2)

Conventional approaches such as logistic regression and gradient boosting have long served as reliable tools for predictive modeling in the clinical domain.(3-5) For example, Framingham Risk Scores quantify the risk of developing cardiovascular conditions, including coronary heart disease, stroke, and heart failure over a period of ten years.(6) However, the continuous advancement of deep learning methods offers the promise of improved prediction accuracy and the ability to extract intricate patterns from complex clinical data.(7)

While the impact of deep learning has been primarily observed in domains that involve unstructured data, such as imaging and natural language processing, its application to structured observational healthcare data has been met with challenges.(7-9) Sparsity, high dimensionality, and heterogeneity have been identified as inherent characteristics of observational healthcare data that limit the efficacy of deep learning methods.(7, 10-12) As a result, conventional prediction methods, such as logistic regression, continue to achieve comparable performance to deep learning methods, despite the complexity of the latter.(7)

However, recent work in deep learning revisits the application to structured tabular data, yielding promising new approaches.(13-15) This study presents a comparison of conventional and deep learning methods to predict three clinically relevant health outcomes using observational healthcare data. Definitions of health outcomes are taken from published clinical articles and include dementia in persons aged 55 – 84, bipolar disorder in patients newly diagnosed with major depressive disorder, and lung cancer in patients aged 45 – 65 who have been cancer-free.(16-18) Moreover, transportability of the models is evaluated across a large network of observational databases through external validation, an aspect not widely studied for deep learning.

# 2 Methods

## 2.1 Source of data

This retrospective study uses structured observational healthcare data from administrative claims and electronic health records (EHR) from the Observational Health Data Sciences and Informatics (OHDSI) collaborative network. Table 1 presents the eleven longitudinal observational databases that were included in this study. The databases include data from the United States, Europe, and Asia-Pacific regions, covering primary and tertiary care center and ranging in population from fewer than 3 million to more than 170 million person records. Further details about the databases are presented in the appendix (pp 2–4). Databases were mapped to the Observational Medical Outcomes Partnership Common Data Model (OMOP CDM).(19) The OMOP CDM provides a standardized data structure and vocabulary, which facilitates sharing and execution of analysis packages across data sites.

**Table 1. Data sources that are in part used for model development and all used for external validation.**

| Database | Acronym | Person count | Usage | Country | Data type | Time period |
|---|---|---|---|---|---|---|
| Integrated Primary Care Information (version N) | IPCI | 2·7M | Development, Validation | Netherlands | EHR (GP) | 01/2006 – 12/2022 |
| Ajou University School of Medicine (version 535) | AUSOM | 2·7M | Development, Validation | South Korea | EHR | 01/1994 – 02/2023 |
| Optum® de-identified Electronic Health Record dataset (version 2541) | Optum EHR | 111·4M | Development, Validation | United States | EHR | 01/2007 – 12/2022 |
| Optum's de-identifed Clinformatics® Data Mart Database (version 2559) | Clinformatics | 97·3M | Development, Validation | United States | Claims | 05/2000 – 03/2023 |
| Stanford Medicine Research Data Repository OMOP (version 24_03_21) | STARR-OMOP | 3·9M | Development, Validation | United States | EHR | 2008 – 2024 |
| Columbia University Irving Medical Center (version 2023q4r2) | CUIMC | 6·1M | Development, Validation | United States | EHR | 1985 – 2023 |
| IQVIA® Disease Analyzer Germany (version 2784) | German DA | 32·8M | Validation | Germany | EHR (GP) | 10/2013 – 09/2023 |
| Japan Medical Data Center (version 2906) | JMDC | 17·7M | Validation | Japan | Claims | 01/2005 – 06/2023 |
| Merative® MarketScan® Multi-State Medicaid (version 2888) | MDCD | 36·0M | Validation | United States | Claims | 01/2000 – 01/2024 |
| Merative® MarketScan® Medicare Supplemental Database (version 2886) | MDCR | 11·3M | Validation | United States | Claims | 01/2000 – 01/2024 |
| Merative® MarketScan® Commercial Claims and Encounters Database (version 2887) | CCAE | 172·3M | Validation | United States | Claims | 01/2000 – 01/2024 |

EHR: electronic health records

GP: general practitioner data included

## 2.2 Prediction problems

A patient-level prediction model quantifies a person's risk of developing a health outcome during a specified time-at-risk period following an index date, using information collected in an observation window prior to index (Figure 1). Health outcomes of interest include:

- **Dementia:** A progressive neurodegenerative disorder characterized by declining cognitive function, including memory loss, impaired thinking, and behavioral changes. Early identification of dementia can facilitate timely intervention, potentially slowing disease progression and improving quality of life.(20) Dementia is predicted in a target cohort of persons aged 55 – 84 during a time-at-risk of 1,825 days as defined in a published clinical article.(18)
- **Bipolar disorder:** A mental health condition marked by recurring episodes of mania or hypomania and depression, significantly impacting mood, energy, and activity levels. A considerable fraction of major depressive disorder patients later receive a corrected diagnosis of bipolar disorder, making early prediction important for effective management.(21-24) The transition to bipolar disorder is predicted in a target cohort of persons diagnosed with major depressive disorder during a time-at-risk of 365 days as defined in a published clinical article.(16)
- **Lung cancer:** The leading cause of cancer mortality in the United States, with a 5-year survival rate of only 22% due to late-stage diagnosis.(25) Despite the proven benefits of screening, uptake is low, and many patients diagnosed with lung cancer do not meet current screening criteria.(26-28) Early detection of lung cancer can significantly improve treatment outcomes and survival rates. Lung cancer is predicted in a target cohort of persons aged 45 – 65 during a time-at-risk of 1,095 days as defined in a published clinical article.(17)

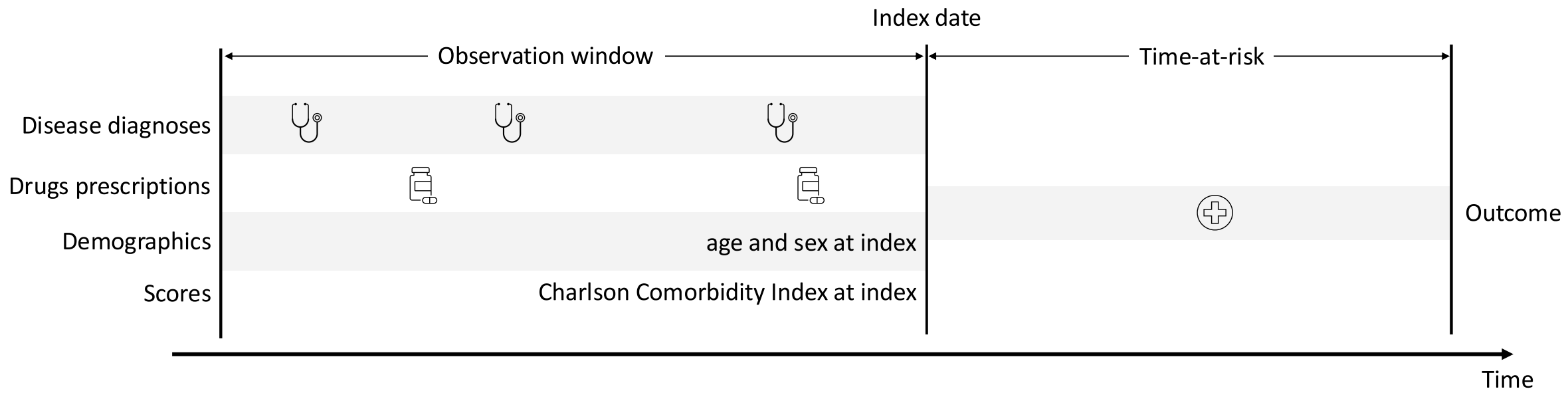

**Figure 1. Onset prediction time windows for a person in a study population.(29)**

For all prediction tasks, a visit record that marks an interaction with a healthcare provider, serves as the index date, allowing for the practical application of the model. To use recent data, but at the same time eliminate pandemic-related confounding effects on healthcare systems and patient behavior, we chose to utilize pre-COVID data right before the pandemic (before 1 January 2020). Given the respective time-at-risk periods this means that the index date for dementia falls into the year of 1 Jan 2014 – 31 Dec 2014, for lung cancer into the year of 1 Jan 2016 – 31 Dec 2016, and for bipolar into the years from 1 Jan 2017 – 31 Dec 2019.

Participants require 365 days of continuous observation time before the index date (excluding the index date), in which candidate predictors are assessed (Figure 1). This relatively short period is consistent with other models in literature and, as opposed to all-time lookback, was also found to have only small impact on discrimination and calibration as all-time lookback can vary strongly across patients.(30) As candidate predictors, we use patient's age, sex, and Charlson Comorbidity Index at the index date. During the observation window we use dichotomized diagnoses and drug prescriptions. Even though this information is recorded at multiple time points, for analysis purposes it is flattened into a tabular format; a necessary step for prediction methods not designed to handle sequence data. Additionally, following empirical recommendations on handling patients lost to follow-up, the study allows participants to exit the cohorts at any time during the time-at-risk period, provided they have at least one day of time-at-risk after the index date.(31)

## 2.3 Prediction methods

Clinical predictive modeling has undergone significant progress over recent decades, evolving from traditional statistical methods to modern machine learning techniques.(3, 5) One of the earliest well-known models is the Framingham Risk Score, developed in the late 1990s using logistic regression to predict ten-year cardiovascular risk.(6) Logistic regression has been widely adopted due to its simplicity, robustness, and interpretability, predicting outcomes by applying a logistic function to a weighted sum of input features.(4, 32) Incorporating L1 regularization enabled effective feature selection and further improved the models' scalability on large datasets, making it optimal for clinical prediction on observational data.(33)

More recently, gradient boosting strategies have demonstrated high efficacy in handling feature interactions and missing values by sequentially combining weak learners, typically decision trees, to minimize errors.(34) Gradient boosting has gained widespread popularity for structured data and is extensively being used in machine learning competitions.(14) Notable implementations include Extreme Gradient Boosting (XGBoost), which for example has been used in a framework for predicting smoking-induced noncommunicable diseases, and in models predicting the mortality of patients with acute kidney injury in intensive care.(35-37)

In contrast to these conventional approaches, neural networks, which serve as the foundation for deep learning, were conceptualized as early as 1958 with the introduction of the perceptron.(38) The field advanced significantly in 1986 with the backpropagation algorithm being introduced, enabling effective training of Multi-Layer Perceptrons (MLPs).(39) Residual Networks (ResNet) further addressed challenges such as the vanishing gradient by incorporating residual connections, allowing for increasingly deeper architectures.(40) ResNets have demonstrated diagnostic performance equivalent to that of health-care professionals in detecting disease from medical imaging.(9) The transformer represents another significant leap in deep learning architecture, particularly for natural language tasks.(41-44) It relies on self-attention mechanisms, enabling efficient parallel processing and improved handling of long-range dependencies.(45)

As the focus shifts back to tabular data, adaptions of the ResNet and Transformer have become available, potentially more suitable for clinical prediction on observational healthcare data.(14) We compare logistic regression, gradient boosting (XGBoost), ResNet, and Transformer to evaluate their efficacy for disease onset prediction across eleven observational databases. (4, 14, 35) The comparison includes internal validation, external validation, and a learning curve analysis to assess scalability with increasing data set size.(46) Model parameters and hyperparameter space for training conventional and deep learning methods are presented in the appendix (p 4). For deep learning methods, this parameter choice is based on previously established recommendations.(14) For L1 regularized logistic regression we use an adaptive search method to automatically tune the degree of regularization.(4) XGBoost uses a grid search in combination with a holdout set to automatically determine its hyperparameters.(2)

## 2.4 Statistical analysis

For internal validation, each dataset is partitioned into a 75% training set and a 25% test set. Model performance on the test set is assessed using the area under the receiver operating characteristic curve (AUROC) for discrimination and the $E_{avg}$ for calibration.(47) AUROC measures the probability that a randomly selected patient with the outcome will have a higher predicted risk than one without.(48) The 95% confidence interval (CI) is computed for each AUROC using stratified bootstrapped replicates.(49) Calibration is typically visualized through plots showing agreement between predicted and observed risk across deciles. To simplify comparison of calibration across models and data sources, the single value metric $E_{avg}$ is used, which indicates the average absolute difference between observed and predicted probabilities.(50, 51) For external validation, performance is evaluated on the full external datasets by applying the models to persons matching the target cohort definitions in the external data sources.

We use Friedman's test to detect ranking differences of the discrimination and calibration performances of the different prediction methods.(52) If the null hypothesis for no difference in ranks between the methods is rejected (p-value less than 0·05), we proceed with a post-hoc test to examine all pairwise differences, controlling for multiplicity.(53) The results are plotted in a critical difference (CD) diagram of the Nemenyi test, which shows the mean ranks of each prediction method.

## 2.5 Role of the funding source

The funding source of this study had no role in study design, data collection, data analysis, data interpretation, or writing and submission of the report.

# 3 Results

The analysis includes several steps to develop and validate clinical prediction models, as outlined in Figure 2. Data are extracted for the outcomes of dementia, bipolar disorder, and lung cancer from eleven databases. Logistic regression, gradient boosting, ResNet, and Transformer models are developed on data from six of these databases and validated across all eleven databases. We evaluate discrimination using AUROC and calibration using $E_{avg}$.

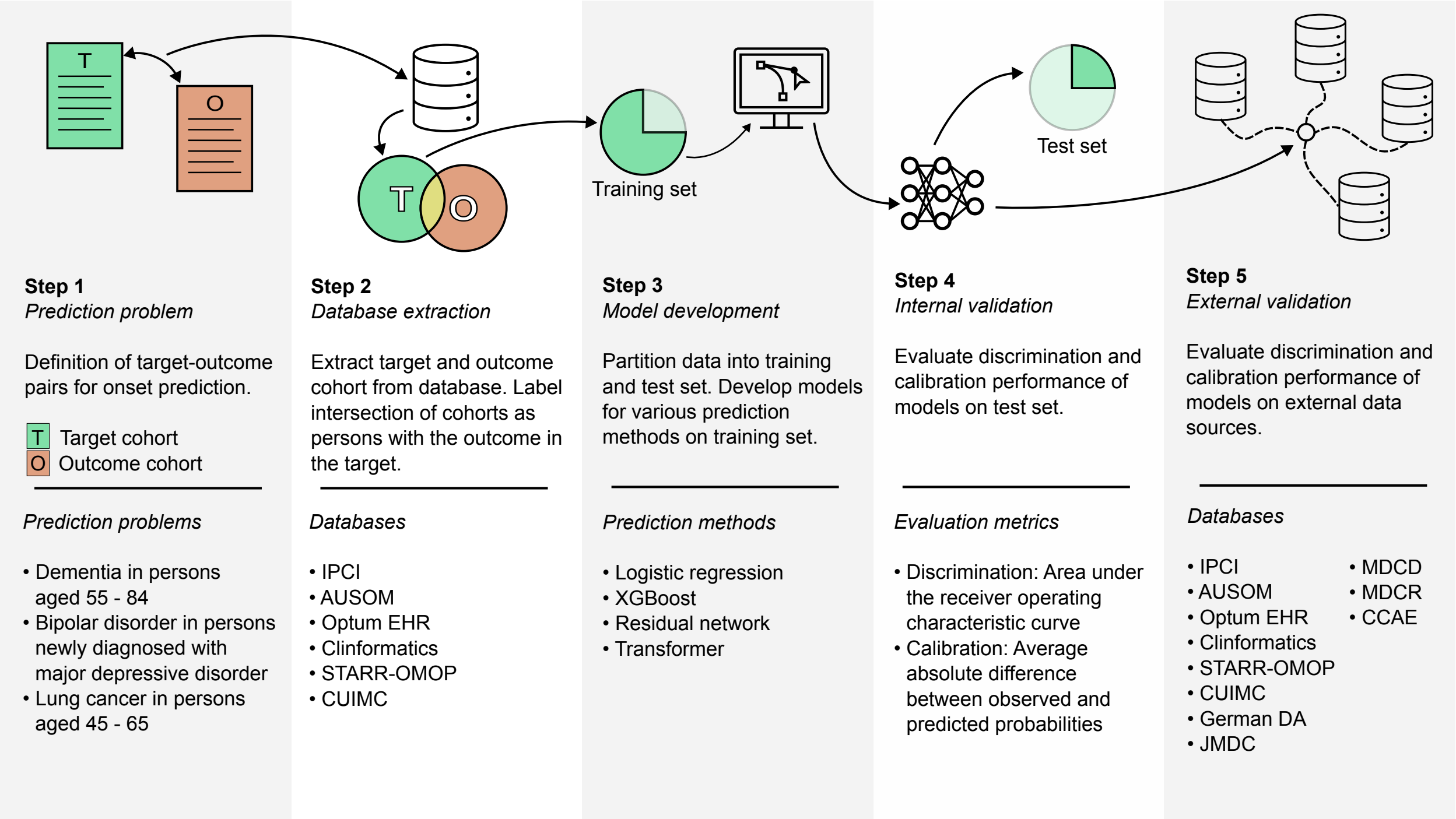

**Figure 2. Study overview.(54)**

## 3.1 Study populations

For dementia prediction, all study populations yield sufficient outcomes for prediction (Table 2). Outcome rates vary between 1·1% in AUSOM and 4·8% in Clinformatics. The two largest data sets from Optum EHR and Clinformatics also provide the largest study populations and most outcomes of dementia. The median time-at-risk for persons in the study populations is the full 1,825 days of the prediction problem definition, except for Clinformatics with a median time-at-risk of 1,746 days. Female representation is notably higher than male representation in Optum EHR, STARR-OMOP, and CUIMC.

**Table 2. Dementia study populations for model development.**

| | IPCI | AUSOM | Optum EHR | Clinformatics | STARR-OMOP | CUIMC |
|---|---|---|---|---|---|---|
| Population, n | 186,767 | 54,723 | 7,811,078 | 2,838,303 | 124,045 | 92,610 |
| Outcomes, n (%) | 3,094 (1·7) | 587 (1·1) | 306,923 (3·9) | 136,018 (4·8) | 2,209 (1·8) | 2,499 (2·7) |
| Median time-at-risk, days (IQR) | 1,825 (443) | 1,825 (645) | 1,825 (575) | 1,746 (1233) | 1,825 (0) | 1,825 (0) |
| Sex | | | | | | |
| Female, n (%) | 100,438 (53·8) | 28,359 (51·8) | 4,524,507 (57·9) | 1,527,140 (53·8) | 72,548 (58·5) | 54,484 (58·8) |
| Male, n (%) | 86,329 (46·2) | 26,364 (48·2) | 3,286,571 (42·1) | 1,311,163 (46·2) | 51,497 (41·5) | 38,117 (41·2) |

For lung cancer prediction, outcome rates are substantially lower despite large study populations (Table 3). The median time-at-risk for persons in the study populations was the full 1,095 days, except for Clinformatics with a median time-at-risk of 916 days. Female representation is notably higher than male representation in Optum EHR, STARR-OMOP and CUIMC.

**Table 3. Lung cancer study population for model development.**

| | IPCI | AUSOM | Optum EHR | Clinformatics | STARR-OMOP | CUIMC |
|---|---|---|---|---|---|---|
| Population | 249,294 | 77,256 | 5,987,928 | 1,435,988 | 92,346 | 45,765 |
| Outcomes, n (%) | 520 (0·2) | 41 (0·1) | 5,937 (0·1) | 1,334 (0·1) | 92 (0·1) | 21 (0·0) |
| Median time-at-risk (IQR) | 1,095 (0) | 1,095 (0) | 1,095 (315) | 916 (751) | 1,095 (0) | 1095 (0) |
| Sex | | | | | | |
| Female, n (%) | 134,228 (53·8) | 35,992 (46·6) | 3,537,268 (59·1) | 738,731 (51·4) | 59,923 (64·9) | 28,590 (62·5) |
| Male, n (%) | 115,066 (46·2) | 41,264 (53·4) | 2,450,660 (40·9) | 697,257 (48·6) | 32,420 (35·1) | 17,171 (37·5) |

For bipolar disorder prediction, the largest study populations are again seen in the largest databases, Optum EHR and Clinformatics. The IPCI and AUSOM cohorts were excluded from the study due to having fewer outcomes than the minimum cell count required for publication. Median time-at-risk was the full 365 days for all cohorts. Female representation is notably higher than male representation across the remaining study populations.

**Table 4. Bipolar disorder study population for model prediction.**

| | IPCI | AUSOM | Optum EHR | Clinformatics | STARR-OMOP | CUIMC |
|---|---|---|---|---|---|---|
| Population | <5 outcomes | <5 outcomes | 842,194 | 398,452 | 16,354 | 10,566 |
| Outcomes, n (%) | | | 12,133 (1·4) | 4,654 (1·2) | 103 (0·6) | 166 (1·6) |
| Median time-at-risk (IQR) | | | 365 (0) | 365 (34) | 365 (0) | 365 (0) |
| Sex | | | | | | |
| Female, n (%) | | | 541,082 (64·2) | 237,183 (59·5) | 11,226 (68·6) | 7,007 (66·3) |
| Male, n (%) | | | 301,112 (35·8) | 161,269 (40·5) | 5,120 (31·3) | 3,555 (33·6) |

## 3.2 Discrimination performance

Internal and external discrimination performance for logistic regression, XGBoost, ResNet, and Transformer for eleven observational databases is presented in Figure 3. Internal validation performance is assessed on the test set partition, whereas external validation performance is assessed on the whole data of each respective validation database and statistically summarized. Models generally perform better during internal validation. For development datasets with few outcomes, AUROC confidence intervals tend to be wider. Complete discrimination results are presented in the appendix (p

5).

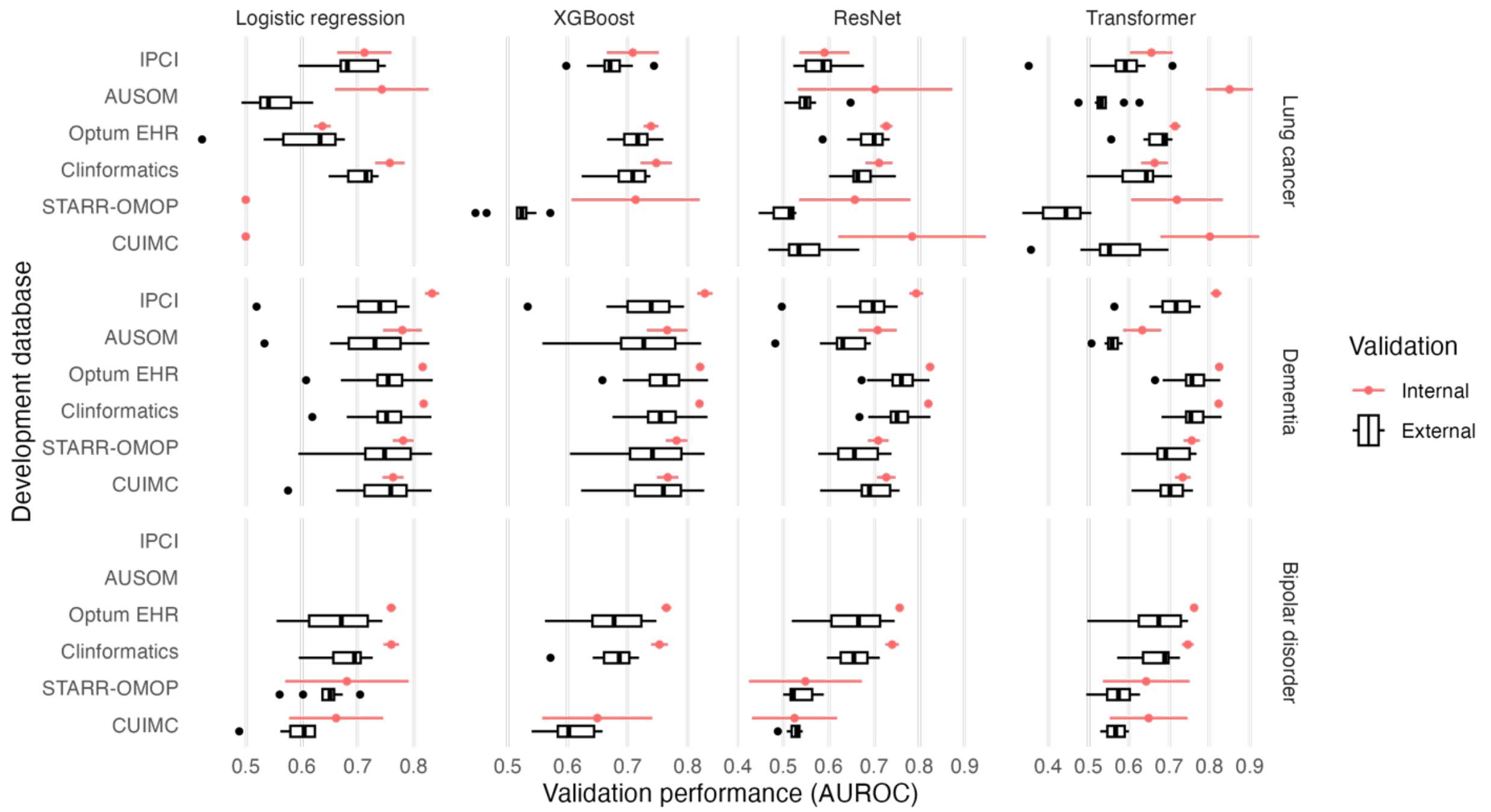

**Figure 3. Discrimination performance (AUROC) for internal validation with 95% CI and for external validation across the respective external databases.**

For discrimination measures across outcomes, databases, internal and external validation, the Friedman test yields a test statistic (Q) of 38·429 with 3 degrees of freedom and rejects the null hypothesis of no difference in ranks (p ≪ 0·0001). This suggests that at least one of the prediction methods differs significantly from the others in discrimination performance. The CD diagram in Figure 4A reveals that XGBoost and Logistic Regression show no significant difference in performance. Similarly, no significant performance difference is observed between Transformer and ResNet. However, conventional models have lower rank compared to deep learning models, indicating a significant performance difference between the methods. Looking only at internal prediction performance, the Friedman test reveals no significant difference between the prediction methods (Q(3) = 6·225, P = 0·10), and, thus, no post-hoc test is performed. However, a significant difference between models shows for external validation performance (Q(3) = 34·555, P ≪ 0·0001), and the CD diagram (Figure 4B) reveals the same ranking as across all discrimination measures. Conventional models developed on the smaller datasets of IPCI, AUSOM, STARR-OMOP, and CUIMC and validated across all data sources except Clinformatics and Optum EHR (Figure 4C) perform better than deep learning models (Q(3) = 40·622, P ≪ 0·0001), while on the larger data sets Clinformatics and Optum EHR conventional and deep learning models perform the same (Q(3) = 1·725, P = 0·63).

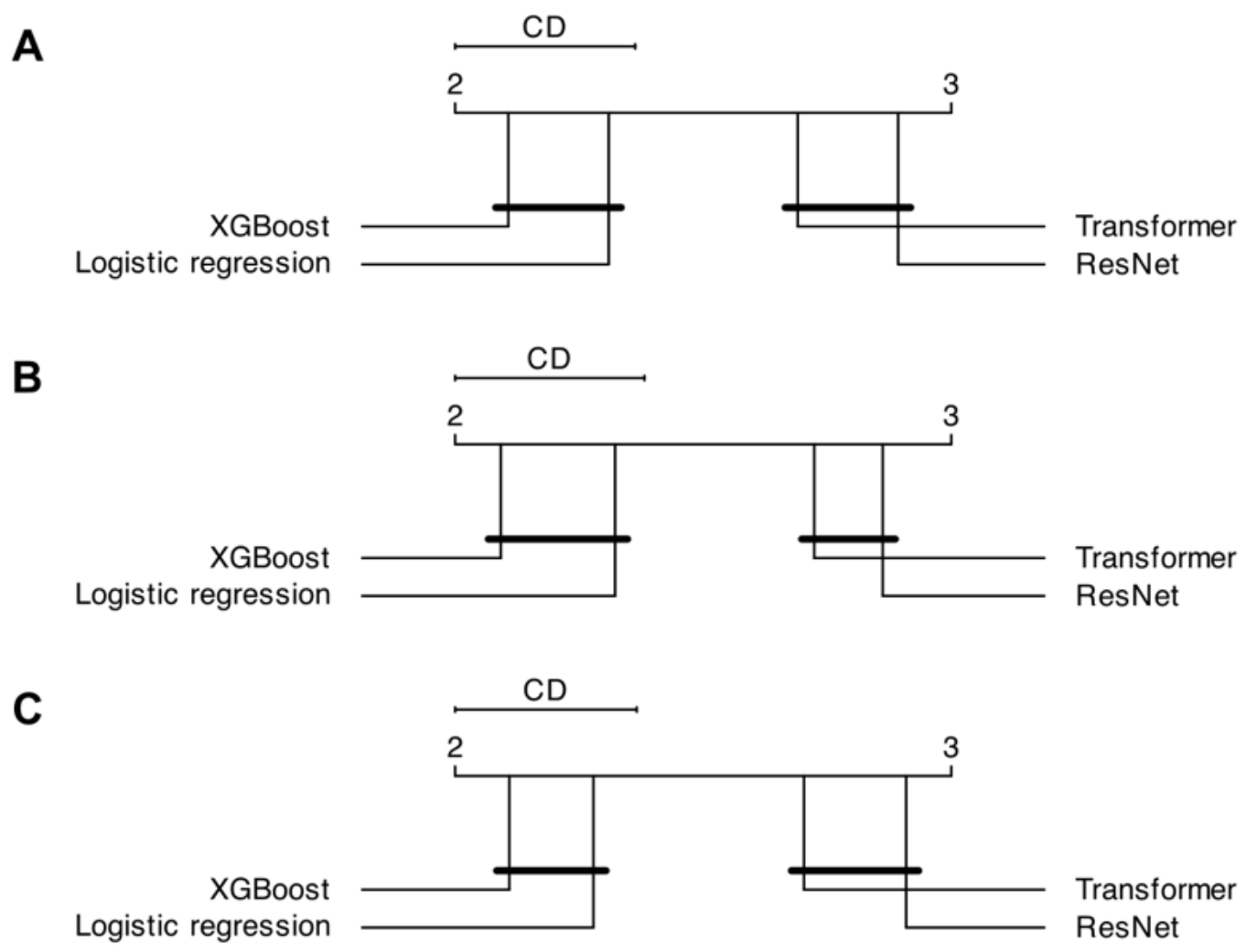

**Figure 4. Ranking of prediction method based on discrimination performance (AUROC) for (A) internally and externally validated models, (B) externally validated models, (C) models developed on small datasets IPCI, AUSOM, STARR-OMOP, and CUIMC and validated across all data sources except Optum EHR and Clinformatics.**

## 3.3 Calibration performance

Calibration performance of the models is evaluated using the $E_{avg}$. Complete calibration results are presented in the appendix (p 6). For measures across outcomes, databases, internal and external validation, the Friedman test yields a test statistic (Q) of 116·79 with 3 degrees of freedom. The p-value is much less than 0·0001, indicating a significant result. The CD diagram in Figure 5A reveals that logistic regression, XGBoost, and Transformer show better calibration than ResNet. The same does not apply to internal validation performance (Q(3) = 29·325, P ≪ 0·0001) where the ResNet is calibrated equally well as the Transformer, but for the external (Q(3) = 95·139, P ≪ 0·0001) validation the ResNet continues to get outperformed (Figure 5B and Figure 5C, respectively). When looking only at models that achieve at least 0·7 AUROC discrimination performance there is no significant difference between them (Q(3) = 4·0621, P = 0·25). For models developed on small datasets IPCI, AUSOM, STARR-OMOP, and CUIMC and validated across all data sources except Clinformatics and Optum EHR the ResNet is outperformed (Q(3) = 118·47, P ≪ 0·0001) as evident in Figure 5D. For models developed and validated on large datasets Clinformatics and Optum EHR (Figure 5E) there is a significant performance difference observed between XGBoost and ResNet (Q(3) = 15·409, P = 0·0015).

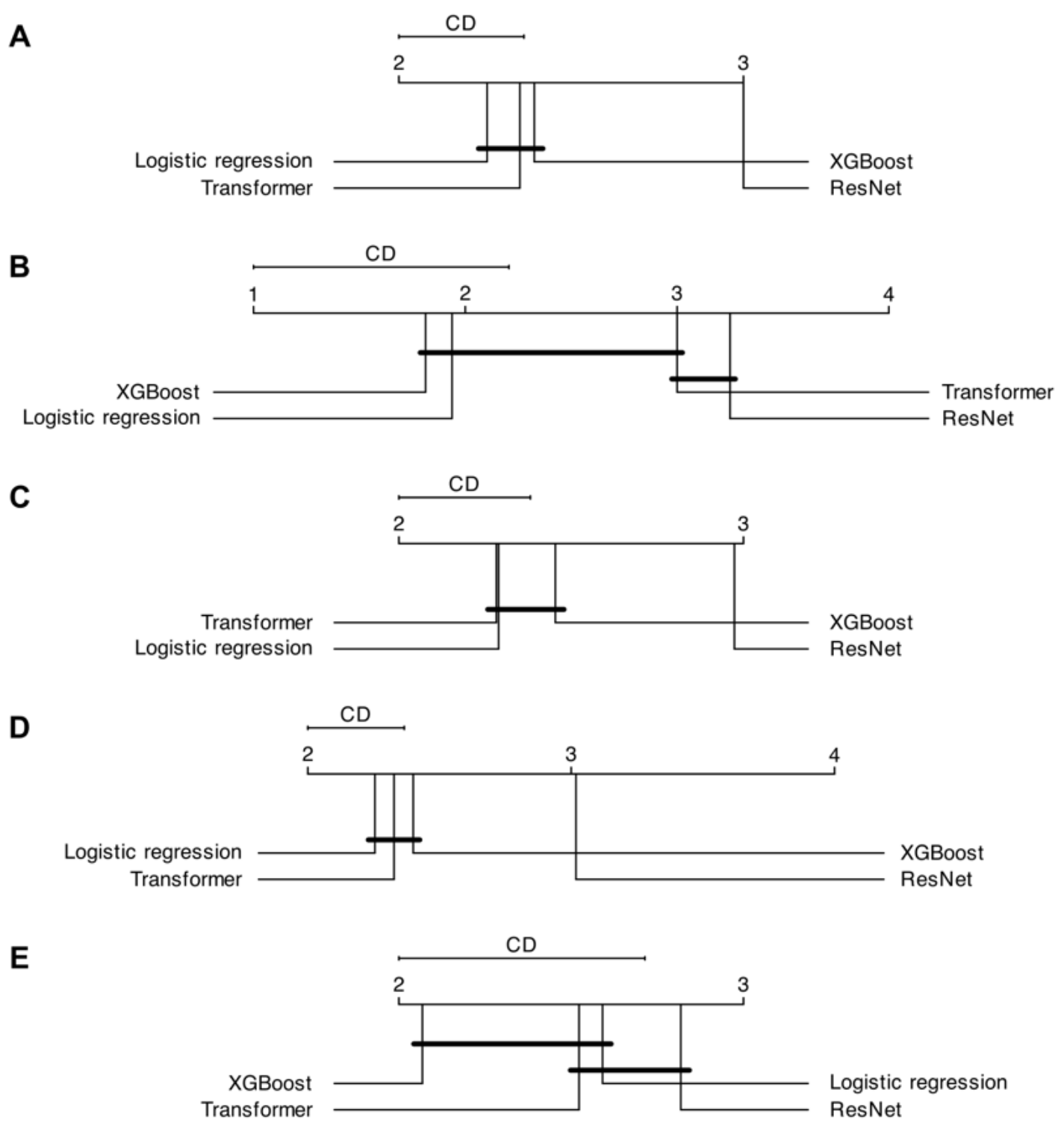

**Figure 5. Ranking of prediction method based on calibration performance ($E_{avg}$) for (A) internally and externally validated models, (B) internally validated models, (C) externally validated models, (D) models developed on small datasets IPCI, AUSOM, STARR-OMOP, and CUIMC and validated across all data sources except Optum HER and Clinformatics, (E) models developed and validated on large datasets Optum EHR and Clinformatics.**

## 3.4 Learning curves

Learning curves illustrate how prediction performance changes with increasingly larger subsets of the training data (Figure 6). Previous work has shown that for logistic regression and clinical prediction tasks with imbalanced classes, the outcome count correlates more closely with performance than training set size.(46) Thus, performance is plotted against outcome count rather than subset size. Given the extensive training times for deep learning models, we reuse the optimal hyperparameters from the model trained on the full training set, rather than performing a new search for each subset.

Learning curves for conventional models require less data to reach the performance plateau. However, with sufficient data performance for all methods appears to converge. ResNet can show a large degree of instability as compared to the other methods, which only diminishes with increasing subset size.

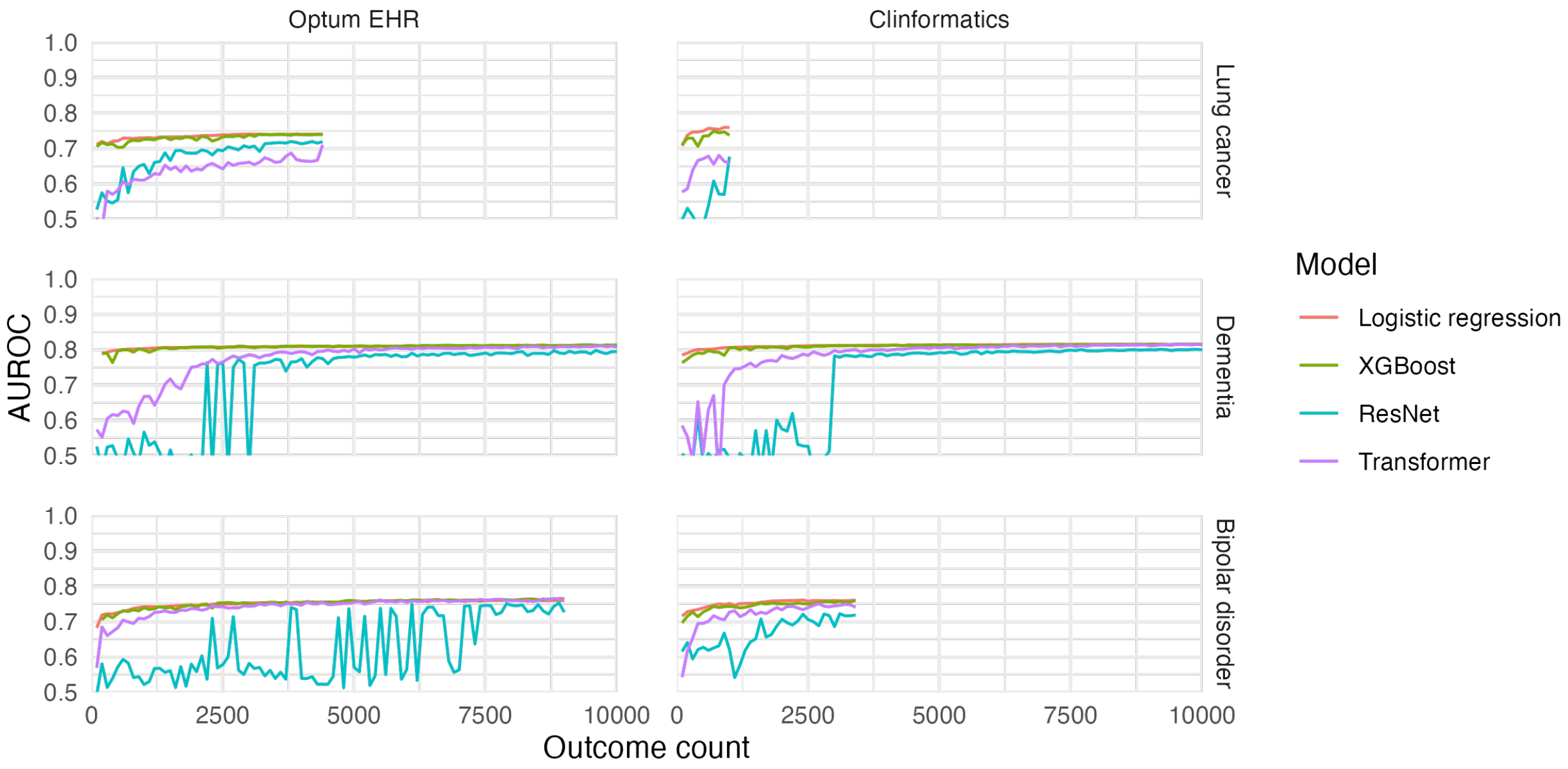

**Figure 6. Discrimination performance (AUROC) on the test set for increasingly larger subsets of the training set.**

# 4 Discussion

In this study, we compared the performance of conventional and deep learning methods for disease onset prediction using observational healthcare data. Our large-scale analysis included logistic regression, XGBoost, ResNet, and Transformer models. We performed internal and external validation across eleven databases in North America, Europe, and Asia, each with different data capture processes and population sizes. Evaluation was performed across three clinical prediction problems with consistent evaluation metrics to ensure robustness and generalizability. Our findings highlight the current capabilities and limitations of deep learning models in the context of structured observational healthcare data.

The study populations reveal considerable differences in cohort size and outcome count, with the largest cohorts observed in Optum EHR and Clinformatics. For bipolar disorder, insufficient data were available for IPCI and AUSOM, which were hence excluded from the analysis. Notably, female representation was higher across the EHR data sets from the United States, which for dementia aligns with existing literature indicating a higher prevalence in women, and in general with the gender difference in utilization of healthcare and preventive care services in the United States.(55-57) Median time-at-risk was the full time-at-risk for all data sets, with the exception of Clinformatics which for dementia is 1,746 days and for lung cancer 916 days. This indicates that few persons are lost to follow up.(31)

Internal and external discrimination performance was evaluated using AUROC across all prediction methods. The CD plot in Figure 4A shows that overall (internal- and externally), conventional methods outperform deep learning methods. This trend is also observed for external validation, indicating limited transportability of the deep learning methods. Interestingly, internal validation shows no statistical difference between methods. Further investigation using learning curves confirms this observation, where in the dementia dataset with the largest cohort sizes, the performance of all prediction methods converges, with the ResNet being slightly outperformed. Analyses for lung cancer and bipolar disorder showed that the learning curves were still in the upwards trending phase and no plateau was reached due to insufficient data. Although learning curves agree with the Friedman test that models perform equally on their development data, deep learning models require substantially larger datasets to reach equivalent performance levels, explaining their lag in smaller datasets. From the learning curves we observe that ResNet shows greater instability in discrimination performance as compared to the other methods. This may suggest that ResNet is more sensitive to hyperparameter choices, as these were derived from the model trained on the full dataset instead of being optimized for each learning curve subset due to lengthy training times. This limitation may also affect the learning curve of the Transformer.

Other notable observations include the absence of several models for which no performance could be measured, likely due to insufficient data or failure to achieve model convergence, and as a result the unavailability of a developed model for external validation. Lack of sufficient data also lead to poor model performance (AUROC close to 0·50) for several models in the lung cancer and bipolar disorder cohorts, whereas dementia results were complete. AUROC values below 0·50 are observed during some external validations indicating systematic misclassification likely caused by overfitting to insufficient training data during model development. Another potential reason for poor external validation performance is database heterogeneity. Although, the conversion to the OMOP CDM provides data standardization and harmonization, limitations apply such as incomplete data capture, variable coding practices, and differences in patient populations across databases. An example of the latter is dementia models consistently performing poorly on CCAE, which can be explained by CCAE not providing the required age range for the target cohort, but only from 55 – 65 years of age. On the other end of the performance range several above average performing models warrant further investigation into potential anomalies. For example, the AUSOM Transformer for lung cancer performs well internally with 0.85 AUROC with wide confidence intervals (0·79 – 0·91) likely driven by a low outcome count of 41, and as a result, does not generalize well to the remaining databases. The Optum EHR Transformer for lung cancer achieves an AUROC of 0.88 at CUIMC. However, the low outcome count of 21 and wide confidence intervals (0·68 – 1·00) warrant cautious interpretation.

Internal calibration performance is good across all models (appendix p 5). Calibration on external databases deteriorates and is worst for dementia, which can be explained by the differences in demographics across databases as age is a driving predictor. Calibration is also found to be excellent for many models that discriminate poorly. We believe that calibration should not be assessed in this case and additionally assess calibration for models that achieve AUROC ≥ 0·70. We present this threshold without specific label, acknowledging that it is somewhat arbitrary and based on digit preference to distinguish from a model with AUROC that predicts not better than chance.(58, 59) Since there is no significant difference found for this subset of models, we conclude that models that perform above 0·70 AUROC are equally well calibrated.

Other insights from this study include the lengthy training times and advanced hardware requirements, such as graphics processing units (GPU), for developing deep learning models. Due to computational constraints on cloud infrastructure with high GPU costs, our GPU hours were limited. Consequently, we prioritized external validation across more databases, as it is less computationally intensive and more cost-effective than model development. These challenges are less pronounced in studies using conventional models and highlights the complexities and costs of deep learning studies. If for a study deep learning models are preferred and given ResNet's poor calibration on small data and instability observed in discrimination-based learning curves, we recommend the use of Transformers if its lengthy training times are acceptable. However, based on the tested architectures, we currently recommend logistic regression or XGBoost over deep learning methods for disease onset prediction on structured observational data, especially for smaller datasets and when external validation is important.

Our finding establishes a critical baseline and highlights the current limitations of deep learning methods when applied to structured observational data. These methods are more complex, but do not achieve better performance than conventional methods. Therefore, it is necessary to better understand the context in which deep learning may offer advantages. Structured observational data likely does not exploit the full capabilities of deep learning methods. Specifically, the removal of temporal information to achieve compatibility with conventional prediction methods is unnecessary for a Transformer model which can effectively use sequence data. Also, the use of potentially more informative features should be considered. For example, measurement data or medical procedures may offer a more comprehensive picture of condition severity. Moreover, pre-training on large databases, as demonstrated by models like Med-BERT, BEHRT, and CEHR-BERT, has shown considerable promise. Pre-training allows models to capture a wide range of medical concepts and patterns, thereby improving generalizability.(60-62)

Our comparison reveals that conventional models remain highly competitive for disease onset prediction using structured observational data. While deep learning holds the potential to excel in more advanced data contexts, conventional prediction methods continue to offer reliable performance and good transportability especially with smaller datasets.

**Contributors**

L.H.J. and E.A.F. lead and C.K. and J.M.R. contributed to the conception and design of the work. L.H.J., C.K., J.M.R, and E.A.F prepared the formal analysis. L.H.J., C.K., J.C., H.M.C., C.P. and E.A.F. accessed, curated, and verified the underlying data. J.A.K. and P.R.R. provided critical feedback on methodology. L.H.J., C.K., J.C., H.M.C., P.D., C.P., and E.A.F. carried out the formal analysis. L.H.J. took the lead in writing the manuscript. All authors provided critical feedback on the manuscript and helped shape the manuscript. All authors read and approved the final manuscript.

**Data sharing**

Data dictionaries defining each field in the set are publicly available as part of the Observational Medical Outcomes Partnership Common Data Model (https://ohdsi.github.io/CommonDataModel/). Requests for analysis results must be formally addressed to L.H.J. The study repository is publicly available at https://github.com/ohdsi-studies/DeepLearningComparison and includes the protocol and study parameters, as well as source code to generate disseminated results. The source code of the analysis framework is publicly available at https://github.com/OHDSI/PatientLevelPrediction and https://github.com/OHDSI/DeepPatientLevelPrediction.

**Declaration of interests**

J.M.R is an employee of Janssen Research & Development and shareholder of Johnson & Johnson. L.H.J., J.A.K., P.R.R., J.M.R., and E.A.F. work for a research group that in the past three years receives/received unconditional research grants from Chiesi, UCB, Amgen, Johnson and Johnson, Innovative Medicines Initiative and the European Medicines Agency. None of these grants result in a conflict of interest to the content of this paper.

# Supplementary material

**Comparison of deep learning and conventional methods for disease onset prediction**

Luis H. John, MSc[1], Chungsoo Kim, PhD[2], Jan A. Kors, PhD[1], Junhyuk Chang, PharmD[3], Hannah Morgan-Cooper, MSc[4], Priya Desai, MSc[4], Chao Pang, PhD[5], Peter R. Rijnbeek, PhD[1], Jenna M. Reps, PhD[1,6], Egill A. Fridgeirsson, PhD[1]

[1]Department of Medical Informatics, Erasmus University Medical Center, Rotterdam, The Netherlands

[2]Section of Cardiovascular Medicine, Department of Internal Medicine, Yale School of Medicine, New Haven, CT, United States

[3]Department of Biomedical Informatics, Ajou University Graduate School of Medicine, Suwon, Republic of Korea

[4]Stanford School of Medicine and Stanford Health Care, Palo Alto, CA, United States

[5]Department of Biomedical Informatics, Columbia University Irving Medical Center, New York, NY, United States

[6]Janssen Research and Development, Titusville, NJ, United States

## Table of contents

## Supplementary Methods

### Source of data

Several databases from the United States, Europe, and Asia-Pacific regions are used for the analysis.

### Integrated Primary Care Information (IPCI)

The Integrated Primary Care Information (IPCI) database is a longitudinal observational database containing routinely collected data from computer-based patient records of a selected group of general practitioners (GP) throughout the Netherlands.(1) IPCI was started in 1992 by the department of Medical Informatics of the Erasmus University Medical Center in Rotterdam. The current database includes patient records from 2006 on, when the size of the database started to increase significantly. In 2016, IPCI was certified as Regional Data Center. Since 2019 the data is also standardized to the OMOP CDM.

### Ajou University School of Medicine (AUSOM)

The Ajou University School of Medicine (AUSOM) database is the EHR database of the Ajou University Medical Center from 1994. Ajou University Medical Center in South Korea is a tertiary teaching hospital with 1,108 beds, 33 medical departments, and 23 operating rooms. The AUSOM database is standardized to the OMOP CDM.

### Optum® de-identified Electronic Health Record dataset (Optum EHR)

Optum's longitudinal EHR repository (Optum EHR) is derived from dozens of healthcare provider organizations in the United States. The data is certified as de-identified by an independent statistical expert following HIPAA statistical de-identification rules and managed according to Optum® customer data use agreements . Clinical, claims and other medical administrative data is obtained from both Inpatient and Ambulatory electronic health records (EHRs), practice management systems and numerous other internal systems. Information is processed, normalized, and standardized across the continuum of care from both acute inpatient stays and outpatient visits. Optum® data elements include demographics, medications prescribed and administered, immunizations, allergies, lab results (including microbiology), vital signs and other observable measurements, clinical and inpatient stay administrative data and coded diagnoses and procedures. In addition, Optum® uses natural language processing (NLP) computing technology to transform critical facts from physician notes into usable datasets. The NLP data provides detailed information regarding signs and symptoms, family history, disease related scores (i.e. RAPID3 for RA, or CHADS2 for stroke risk), genetic testing, medication changes, and physician rationale behind prescribing decisions that might never be recorded in the EHR.

### Optum's de-identifed Clinformatics® Data Mart Database (Clinformatics)

Optum's Clinformatics® Data Mart (Clinformatics) is derived from a database of administrative health claims for members of large commercial and Medicare Advantage health plans. Clinformatics® Data Mart is statistically de-identified under the Expert Determination method consistent with HIPAA and managed according to Optum® customer data use agreements. Clinformatics administrative claims submitted for payment by providers and pharmacies are verified, adjudicated and de-identified prior to inclusion. This data, including patient-level enrollment information, is derived from claims submitted for all medical and pharmacy health care services with information related to health care costs and resource utilization. The population is geographically diverse, spanning all 50 states.

### Stanford Medicine Research Data Repository OMOP (STARR-OMOP)

The Stanford Medicine Research Data Repository OMOP (STARR-OMOP) is Stanford's second-generation clinical data warehouse dataset designed to enhance access to healthcare data for research purposes. Launched in 2019, STARR-OMOP contains electronic health records data from Stanford Health Care and Stanford Children's Hospital, the TriValley Hospital, and associated clinics for around four million patients from 2008. The dataset is refreshed monthly and contains demographics, labs, diagnoses, drugs, and procedure information, as well as clinical notes. Several flowsheet fields are also mapped to the OMOP measurements table, including vitals such as blood pressure, oxygen level, heart rate, respiratory rate, measurements from the Sequential Organ Failure Assessment (SOFA) score, etc. The structured and unstructured data is anonymized using a combination of Safe Harbor and other techniques. Based on guidelines from our University Privacy Office (UPO), our location data contains zip5 data for over 77% of the population. Our data and ETL processes are entirely hosted on the cloud.

**Columbia University Irving Medical Center (CUIMC)**

The Columbia University Irving Medical Center (CUIMC) database holds electronic health records for more than six million patients, with data collection starting in 1985. CUIMC, located in the northeast US, is a quaternary care center that provides primary care in northern Manhattan and nearby areas, covering both inpatient and outpatient services. The database encompasses a wide array of data types, including patient demographics, visit details for inpatient and outpatient care, conditions (billing diagnoses and problem lists), medications (outpatient prescriptions and inpatient medication orders and administrations), medical devices, clinical measurements (such as laboratory tests and vital signs), and other clinical observations like symptoms. The data is sourced from a variety of systems, including current and former electronic health record (EHR) systems such as the homegrown Clinical Information System, WebCIS, Allscripts Sunrise Clinical Manager, Allscripts TouchWorks, and Epic Systems. It also includes data from administrative systems like IBM PCS-ADS, Eagle Registration, IDX Systems, and Epic Systems, as well as ancillary systems such as the homegrown Laboratory Information System (LIS), Sunquest, and Cerner Laboratory. The CUIMC data has been standardized to the Observational Medical Outcomes Partnership Common Data Model (OMOP CDM).

**IQVIA® Disease Analyzer Germany (German DA)**

IQVIA® Disease Analyzer Germany (German DA) is a longitudinal patient database providing anonymized information from continuing physician and patient interaction on consultations, diagnoses and treatment within Primary Care. It contains a data from approximately 2,500 office based doctors in Germany. The contents of the database document the management of patients by General Practitioners as well as some specialists and include comprehensive records of diagnosis information; the management of the diagnosis, be it prescription issues, hospital admission or other tertiary care; specialist referrals; laboratory test results and administrative activities. All the events are date stamped, with diagnosis/note/test information collected. Prescriptions, issued by GPs using either the generic substance or drug name are captured exactly as written, including information on indication, dose, strength and dosage instruction and cost.

**Japan Medical Data Center (JMDC)**

Japan Medical Data Center (JMDC) database consists of data from more than 250 Health Insurance Associations covering workers aged less than 75 and their dependents. The proportion who are younger than 66 years old in JMDC is approximately the same as the proportion in the whole nation. JMDC data includes data on membership status of the insured people and claims data provided by insurers under contract. Claims data are derived from monthly claims issued by clinics, hospitals and community pharmacies. The size of JMDC population is about 10% of people in the whole nation.

**Merative® MarketScan® Multi-State Medicaid Database (MDCD)**

The Merative® MarketScan® Multi-State Medicaid Database (MDCD) reflects the healthcare service use of individuals covered by Medicaid programs in numerous geographically dispersed states. The database contains the pooled healthcare experience of Medicaid enrollees, covered under fee-for-service and managed care plans. It includes records of inpatient services, inpatient admissions, outpatient services, and prescription drug claims, as well as information on long-term care. Data on eligibility and service and provider type are also included. In addition to standard demographic variables such as age and gender, the database includes variables such as federal aid category (income based, disability, Temporary Assistance for Needy Families) and race.

**Merative® MarketScan® Medicare Supplemental Database (MDCR)**

The Merative® MarketScan® Medicare Supplemental Database (MDCR) represents the health services of retirees in the United States with Medicare supplemental coverage through employer-sponsored plans. This database contains primarily fee-for-service plans and includes health insurance claims across the continuum of care (e.g. inpatient, outpatient and outpatient pharmacy).

### Merative® MarketScan® Commercial Claims and Encounters Database (CCAE)

The Merative® MarketScan® Commercial Claims and Encounters Database (CCAE) includes health insurance claims across the continuum of care (e.g. inpatient, outpatient, outpatient pharmacy, carve-out behavioral healthcare) as well as enrollment data from large employers and health plans across the United States who provide private healthcare coverage for employees, their spouses, and dependents. This administrative claims database includes a variety of fee- for-service, preferred provider organizations, and capitated health plans.

### Prediction methods

The hyperparameter space that is explored during training is presented in Table S1, Table S2, Table S3, and Table S4 for logistic regression, XGBoost, ResNet, and Transformer, respectively. For developing ResNet and Transformer models we follow the common approach of sampling 100 combinations from the hyperparameter space to reduce computational costs.(2)

Table S1. Hyperparameter space of logistic regression models.

| **Hyperparameter** | **Values** |
|---|---|
| Prior distribution starting variance (lower limit) | 0.01 |
| Lower prior variance limit for grid-search | 0.01 |
| Upper prior variance limit for grid-search | 20 |

Table S2. Hyperparameter space of XGBoost models.

| **Hyperparameter** | **Values** |
|---|---|
| Number of trees | 100, 300 |
| Maximum depth of each tree | 4, 6, 8 |
| Boosting learning rate | 0.05, 0.1, 0.3 |

Table S3. Hyperparameter space of ResNet models.

| **Hyperparameter** | **Values** |
|---|---|
| Size of embedding layer | 64, 128, 256, 512 |
| Number of layers | 1, 2, 3, 4, 5, 6, 7, 8 |
| Number of neurons in each default layer | 64, 128, 256, 512, 1024 |
| Factor to grow the number of neurons in each residual layer | 1, 2, 3, 4 |
| Dropout after first linear layer in residual layer | 0.00, 0.05, 0.10, 0.15, 0.20, 0.25, 0.30 |
| Dropout after last linear layer in residual layer | 0.00, 0.05, 0.10, 0.15, 0.20, 0.25, 0.30 |

Table S4. Hyperparameter space of Transformer models.

| **Hyperparameter** | **Values** |
|---|---|
| Number of transformer blocks | 2, 3, 4 |
| Embedding dimensions | 64, 128, 256, 512 |
| Number of attention heads | 2, 4, 8 |
| Dropout for attention | 0.00, 0.05, 0.10, 0.15, 0.20, 0.25, 0.30 |
| Dropout for feedforward block | 0.00, 0.05, 0.10, 0.15, 0.20, 0.25, 0.30 |
| Dropout for residual connections | 0.00, 0.05, 0.10, 0.15, 0.20, 0.25, 0.30 |
| Dimension of the feedforward block as a ratio of the embedding dimensions | 0.75 |

# Supplementary Figures

## Discrimination performance

Figure S1 presents the internal and external discrimination performances of all developed models using the area under the receiver operating characteristic curve (AUROC). The 95% confidence intervals (CI) is provided alongside these results.

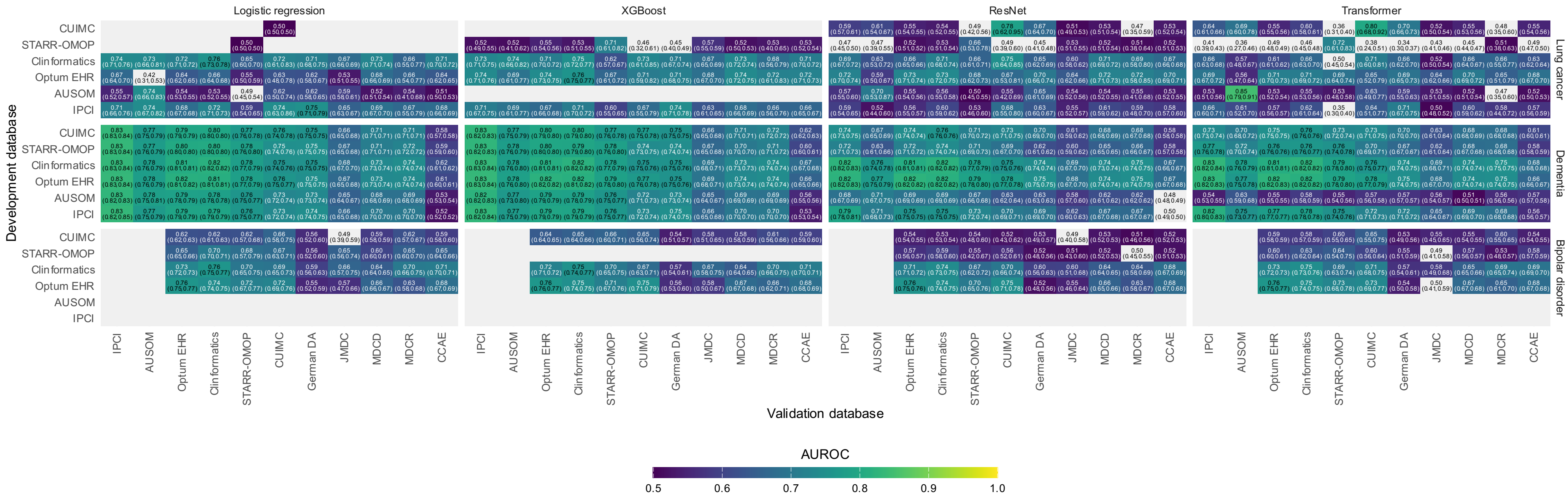

Figure S1. Internal and external discrimination performance (AUROC) with 95% CI across prediction methods and prediction problems.

## Calibration performance

Figure S2 presents the internal and external calibration performances of all developed models using the average absolute difference between observed and predicted probabilities ($E_{avg}$).

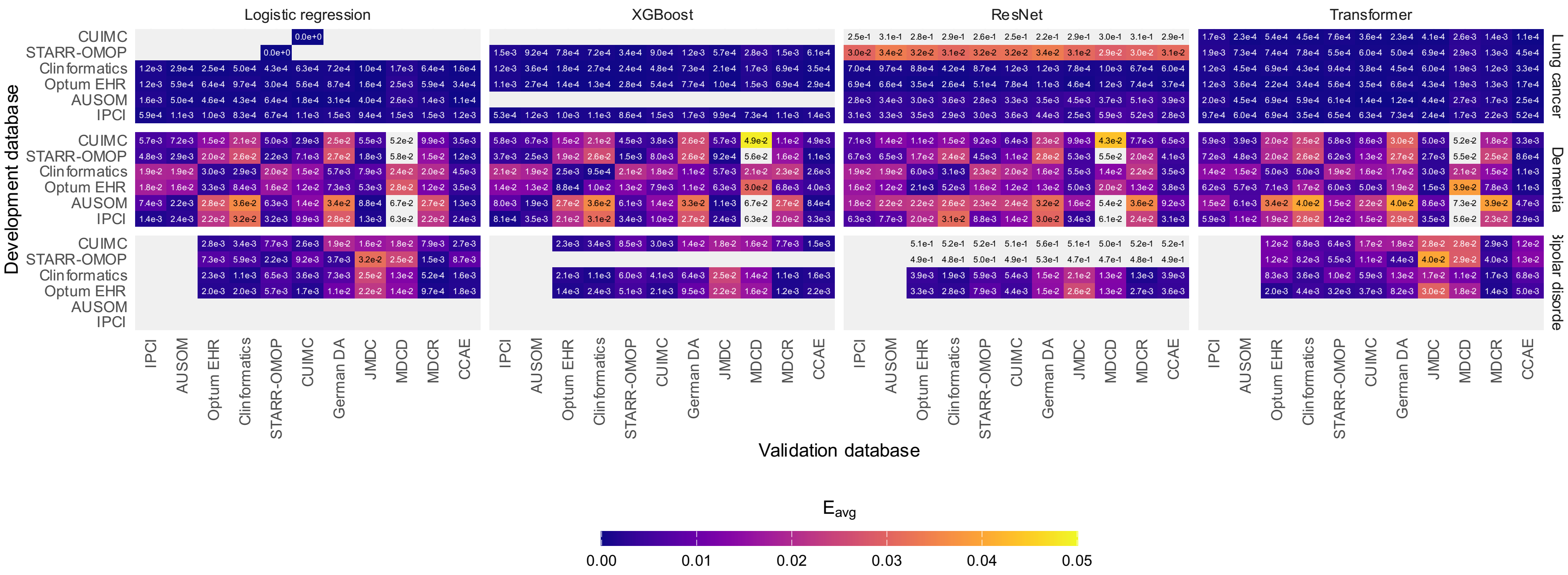

Figure S2. Internal and external calibration performance ($E_{avg}$) across prediction methods and prediction problems.